\documentclass{article}
\pdfoutput=1

\usepackage[nonatbib, preprint]{neurips_2020}

\usepackage[utf8]{inputenc} 
\usepackage[T1]{fontenc}    
\usepackage{hyperref}       
\usepackage{url}            
\usepackage{booktabs}       
\usepackage{amsfonts}       
\usepackage{nicefrac}       
\usepackage{microtype}      
\usepackage{graphicx}

\usepackage{amsmath,amssymb}

\newcommand{\ie}{\textit{i}.\textit{e}., }

\newcommand{\wrt}{\textit{w}.\textit{r}.\textit{t}. }

\newcommand{\V}{\mathcal{V}}
\newcommand{\E}{\mathcal{E}}
\newcommand{\U}{\mathcal{U}}
\newcommand{\Proba}{\mathcal{P}}
\newcommand{\F}{\mathcal{F}}
\newcommand{\M}{\mathcal{M}}
\newcommand{\G}{\mathcal{G}}
\newcommand{\GM}{\mathcal{G}_\mathcal{M}}
\newcommand{\expectation}[2][]{\mathbb{E}_{#1}\left[#2\right]}

\newcommand\indep{\protect\mathpalette{\protect\independenT}{\perp}}
\def\independenT#1#2{\mathrel{\rlap{$#1#2$}\mkern2mu{#1#2}}}

\title{Causal Inference with Deep Causal Graphs}

\author{%
  Álvaro Parafita \\
  Universitat de Barcelona \\
  Barcelona, Spain \\
  \texttt{parafita.alvaro@ub.edu}
  \And
  Jordi Vitrià \\
  Universitat de Barcelona \\
  Barcelona, Spain \\
  \texttt{jordi.vitria@ub.edu}
}

\begin{document}

\maketitle

\begin{abstract}
    Parametric causal modelling techniques rarely provide functionality for counterfactual estimation, often at the expense of modelling complexity. Since causal estimations depend on the family of functions used to model the data, simplistic models could entail imprecise characterizations of the generative mechanism, and, consequently, unreliable results. This limits their applicability to real-life datasets, with non-linear relationships and high interaction between variables. We propose Deep Causal Graphs, an abstract specification of the required functionality for a neural network to model causal distributions, and provide a model that satisfies this contract: Normalizing Causal Flows. We demonstrate its expressive power in modelling complex interactions and showcase applications of the method to machine learning explainability and fairness, using true causal counterfactuals.
\end{abstract}

\section{Introduction}
\label{sec:introduction}

Meaningful scientific inquiry searches for explanations of phenomena, not only descriptions of the data. The objective is to understand why something happened or how to influence an outcome; the mechanism that generates the data, not the data itself. This is the realm of \textbf{causal inference} \cite{pearl2009causality}, where interventional queries can be answered, with applications to artificial intelligence, epidemiology, social sciences or business. As an example, in a loan-approval setting, we could ask what features have the most effect in a refusal (intervention), or why a specific loan was rejected and how that outcome could be changed (counterfactual). Fairness is also related, since we could study the impact of protected variables (\ie gender, race) on a decision system.

There are two elements required for causal estimation: a \textbf{causal graph}, that specifies ordered pairs of variables with a causal relationship ($\textit{A} \rightarrow \textit{B}$, meaning \textit{A} causes \textit{B}), and a \textbf{causal model}, that represents these relationships in a functional form. The former can be derived from domain knowledge and structure learning algorithms (see \cite[chapter~7]{peters2017elements}), and is still an ongoing field of research. This work focuses on the latter, how to model these relationships.

Causal models are usually defined in the form of Structural Equations Models (SEM), where each direct relationship between \textit{causes} and \textit{effects} is modelled by an equation of the form $\textit{effects} = f(\textit{causes})$. Our objective with such a model is the computation of causal expressions, such as the ones mentioned above. Although it is possible, by means of do-calculus \cite{pearl2009causality}, to derive estimands based on the joint distribution of observed variables, these expressions tend to be intractable. Parametric families of SEM functions avoid this issue. However, the use of a parametric function $f$ imposes certain requirements. For example, if we want to compute counterfactuals, having a computable posterior of exogenous noise signals given evidence is mandatory. It is for this reason that SEMs tend to use constrained functional forms: linear equations (possibly with non-linear basis functions) or post-nonlinear models \cite{zhang2012identifiability}, among others. In spite of the advantages of these simple expressions, they lack in flexibility to model complex distributions.

The contribution of this paper is twofold. On the one hand, we propose the \textbf{Deep Causal Graph} (DCG), an implementation contract for Neural Networks to model causal relationships. Such a model allows sampling from the data distribution and from any intervened distributions, computing the log-likelihood of any data point, and, more importantly, estimate interventional and counterfactual queries. As far as we know, there are no truly causal counterfactual estimators other than the aforementioned constrained SEMs. We also provide a model that fits the DCG framework, {\bf Normalizing Causal Flows} (NCF), which leverage the fitting capabilities of Normalizing Flows to model complex causal relationships. 

On the other hand, we showcase applications of our method to the fields of \textbf{black-box counterfactual explainability and fairness}. Specifically, for a given black-box decision system, we are capable of answering the following questions: what is the \textit{causal} effect of an input variable on the outcome; to what extent a feature is responsible for a particular decision and how could we modify it by intervening on the input variables; whether the decision is counterfactually fair, and how to train the system to be fair. 

We provide a complete PyTorch library for modelling causal graphs using our techniques, which includes all DCG models mentioned in this paper and functionality for running experiments on the suggested applications. The code can be found in the supplementary material and soon as an open-source library on Github.

This paper is organized as follows. We review related work in section \ref{sec:related}. Section \ref{sec:background} establishes the required definitions and background knowledge in Structural Equation Models, interventions and counterfactuals. Section \ref{sec:method} defines the Deep Causal Graph, along with the algorithms to perform causal inference. Afterwards, we define Normalizing Causal Flows and include a final subsection on practical considerations. We showcase our approach flexibility in modelling complex datasets in section \ref{sec:experiments}, and demonstrate applications of our model in section \ref{sec:applications}. Finally, we discuss further extensions in section \ref{sec:further} and conclude with section \ref{sec:conclusion}.

\section{Related work}
\label{sec:related}

Structural Equation Models (SEM), according to Pearl \cite{pearl2009causality}, can be dated back to 1921 \cite{wright1921correlation}. Named \textit{path coefficients} at that time, these SEMs represented each node as a linear function of its parents. The approach was extended through basis functions, to provide non-linearity and interaction between input variables (\ie \cite{schumacker1998interaction, lee2002maximum}). Additionally, a non-linearity can be applied to the linear expression, \ie \cite{hoyer2009nonlinear} or use the post-nonlinear causal model \cite{zhang2012identifiability}, both also employed for structure learning. An alternative view, related to our approach, can be found in \cite[chapter~5]{koller2009probabilistic}, where each variable is modelled as a Conditional Probability Distribution (CPD), the distribution of that variable conditioned on its parents. These CPDs can take many forms, from tabular representations of all parameters in the assumed distribution, to Generalized Linear Models.

The proposed Deep Causal Graph further extends SEMs and increases their representation capacity by leveraging the expressive power of deep neural networks. From this point of view, it is directly related to two previous works. CausalGAN \cite{kocaoglu2017causalgan} represented each random variable as a neural network with their parent's values as the input. Distributional Causal Nodes \cite{parafita2019explaining} extended this idea by assuming a known parametric probability distribution for each node. Both were used for computer vision applications, working on descriptors of a picture. However, while the former is trained with adversarial strategies, the latter is able to use Maximum Likelihood Estimation. Our approach is based on the latter, avoiding the distributional assumptions by using Normalizing Flows \cite{papamakarios2019normalizing}.

In terms of applications, Deep Causal Graphs are specially suited to counterfactual explanations. Counterfactual reasoning has been proposed as an important ingredient for explainability and fairness analysis \cite{wachter2017counterfactual, goyal2019counterfactual, mothilal2019explaining}, but in all these frameworks, counterfactuals are understood as samples of the observational distribution, with minimal alterations in the input, that change a black-box prediction. This definition, however, does not take into account the causal effects of these alterations on the rest of the variables, therefore providing non-actionable explanations. Our model does work with intervened distributions, therefore circumventing this issue. Additionally, it allows a practical implementation of Counterfactual Fairness \cite{kusner2017counterfactual}, which ensures that any intervention of protected variables does not entail an effect on the target variable. 

\clearpage
\section{Background}
\label{sec:background}

We define a \textbf{Structural Equation Model} (SEM) as the tuple $\M = (\V, \E, \U, \Proba_\E, \Proba_\U, \F)$, where:

\begin{enumerate}
\item $\V = \{V_1, \dots, V_K\}$ is the set of observable random variables.
\item $\E = \{E_1, \dots, E_K\}$ is the set of exogenous noise variables, one for each $V_k$.
\item $\emptyset \subseteq \U \subseteq \{U_{\{k, l\}}\}_{k, l = 1..K}$ is the set of latent (non-observable) confounder variables $U_{\{k, l\}}$ that explain unobserved common causes between $V_k$ and $V_l$.
\item $\Proba_\E$ and $\Proba_\U$ are the prior distributions for all non-observable variables. 
\item $\F = \{f_k = f_k(Pa_k, U_{\{k, .\}}, E_k)\}_{k=1..K}$ are the functional relationships\footnote{Note that, although $f_k$ is deterministic, the effect of $E_k$ makes it stochastic \wrt $Pa_k$, $U_{\{k, .\}}$.} 
$V_k = f_k(.)$ between a node $V_k$, its observable parent set $Pa_k \subsetneq \V$, its parent latent variables $U_{\{k, .\}}$ (if any) and its corresponding exogenous noise signal $E_k$. 
\end{enumerate}

$\F$ implicitly defines a directed graph $\G_{\M} = (\V \cup \E \cup \U, E)$ where $E$, its edges, are defined by all input-output relationships in $\F$: $E = \bigcup_{k=1..K} \{(V, V_k) \mid V \in Pa_k\} \cup \{(U, V_k) \mid U \in U_{\{k, .\}} \} \cup \{(E_k, V_k)\}$. Any directed edge in the graph represents a causal dependency between source/cause and target/effect. There is only one requirement: $\G_{\M}$ must be a Directed Acyclic Graph (DAG), meaning, it contains no directed cycles. From now on, we assume that the graph $\GM$ is given and we will focus on learning the actual functional relationships for each of the observable nodes, $\F$.

\subsection{Sampling and log-likelihood}

There are several operations that can be carried out with a SEM. On the one hand, we can sample from the observed distribution (the joint distribution of the variables in $\V$) by: 1) sampling from $\E$ and $\U$, using their respective priors $\Proba_\E$ and $\Proba_\U$, and 2) computing values for $\V$ by following a topological order of the graph (all parents come before their children) using the functions in $\F$. 

On the other hand, we can compute the log-likelihood of any sample by using the general product rule of probability and the conditional independencies entailed by the graph $\GM$ (\textit{d-separability}, see \cite{pearl2009causality} or \cite{koller2009probabilistic}). For the case with no latent confounders ($\U = \emptyset$), assuming that the variables in $\V$ are in a topological order of the graph, then $(V_k \indep V_{<k} \mid Pa_k),\, \forall k=1..K$, where $V_{<k} = \{V_1, \dots, V_{k-1}\}$. Therefore, since $Pa_k \subseteq V_{<k}$, $\log p(v_1, \dots, v_K) = \sum_{k=1..K} \log p(v_k \mid v_{<i}) = \sum_{k=1..K} \log p(v_k \mid pa_k)$. Depending on the choice of functions $\F$, we can compute these individual conditional probabilities and, as a result, the joint log-likelihood of a sample. The case with latent confounders is discussed in the following sections.

\subsection{Interventions and counterfactuals}

The former operations work on the observational model of the data. Causality modelling allows another kind of operation, the intervention, that alters the distribution represented by the model. Specifically, a constant-value intervention, normally represented by $\textrm{do}(X=x)$, means replacing the generative function $X = f_X(.)$ by an assignment $X = x$. This constant value $x$ comes predefined by the intervention and does not depend on the parents of $X$. Therefore, the intervened SEM replaces $f_X$ by this assignment and the corresponding intervened graph is the subgraph where all edges pointing at $X$ are removed. This subgraph encodes a different probability distribution, the \textbf{intervened distribution}, from which we can sample and compute log-likelihoods like before. 

The final ingredient in our theory is the \textbf{counterfactual}. Given a certain sample $v$ and an intervention $\textrm{do}(X=x)$, a counterfactual is the result of an hypothetical experiment in the past, the answer to what would have happened to the values of our variables in $\V$ had we intervened on $X$ by assigning value $x$. In other words, counterfactual expressions are of the form $p(V' \mid v, \textrm{do}(X=x))$, with $V'$ the counterfactual target variables of study. Pearl \cite{pearl2009causality} defines counterfactuals as a three-step process: \textbf{abduction}, compute the posterior distribution\footnote{We refer to these new distributions as the \textit{abducted priors}.} of the latent variables $\E$ and $\U$ conditioned on evidence $v$, $p(\E, \U \mid v)$; \textbf{intervention}, apply the desired intervention $\textrm{do}(X=x)$; \textbf{prediction}, compute the required prediction in the intervened, counterfactual model $\widehat{\M}$ defined by the abducted priors and the modified set of functions $\widehat{\F}$, where $f_X$ has been replaced by the assignment $X=x$.

The main hindrance to the implementation of SEMs are these abducted noise priors. A linear SEM (defined by using linear equations in $\F$) computes them by inversion of each $f_X$ using basic algebraic rules. Adding an invertible non-linearity to this linear model is also possible. However, the use of more complex expressions hinders the computation of this posterior. This could explain why not much work deals with applying neural networks (a universal approximator) to causality, specifically to counterfactual estimation, in contrast with our approach, which does not suffer from this problem.

\section{Method}
\label{sec:method}

\subsection{Deep Causal Graph}

A Deep Causal Graph (DCG) is an abstract specification of the required functionality for a Deep Neural Network to work with causal queries. The only assumption is that $p(v) > 0$ for all $v$ in the domain of $\V$, which means, all possible configurations of the graph's variables are possible, no matter how unlikely. In this subsection, we specify the characteristics of the model and the associated algorithms based on this abstract specification.

Firstly, a DCG models the SEM described in section \ref{sec:background}. Given a SEM $\M = (\V, \E, \U, \Proba_\E, \Proba_\U, \F)$, we represent each random variable in $\V$ as a subcomponent of the DCG, called the Deep Causal Unit (DCU). Each DCU requires three operations, which may or may not call for a neural network to implement them. These operations are:

\begin{itemize}
    \item $\textbf{\textit{sample}}(\textbf{parents})$: sample a new realization of the variable, given its parents values.
    \item $\textbf{\textit{loglk}}(\textbf{sample}, \textbf{parents})$: compute the log-likelihood of the sample, given its parents values. This operation is required to be \textbf{differentiable} with respect to its parents.
    \item $\textbf{\textit{abduct}}(\textbf{sample}, \textbf{parents})$: given a sample and its parents, compute the noise posterior.
\end{itemize}

The first observation is that, since Neural Networks are deterministic and nodes are stochastic (they model a random variable) we need a source of randomness to properly model these functions. That is the role of the exogenous noise signals in $\E$, one for each $V_k$ in $\V$. As an example, consider a variable/node $X$ modelled with a univariate Gaussian distribution of parameters $\mu$ and $\sigma$. Then, $\textbf{\textit{sample}}$ could be implemented by taking the corresponding signal $E_X$, assumed with a prior $p(E_X) \sim \mathcal{N}(0, 1)$ and then, given a realization $\varepsilon_x \sim p(E_X)$, compute $x = \mu + \sigma \cdot \varepsilon_x$. Although our $\textbf{\textit{sample}}$ operation is deterministic, its result is stochastic due to $E_X$. 

Given these three operations per variable-node, we can perform inference across the graph. Assuming nodes in topological order, sampling consists of iteratively applying each node's $\textbf{\textit{sample}}$ operation, passing parent values if required. Interventions are performed by replacing the \textit{sample} operation by a simple assignment (the intervened value). Now, given a sample $v = (v_1, \dots, v_K)$ (only considering observable nodes in $\V$), computing its $\textbf{\textit{loglk}}$ consists of applying the general product rule and the independencies defined by the graph (as mentioned in section \ref{sec:background}): $\log p(v) = \sum_{k=1..K} \log p(v_k \mid pa_k)$. Note that when using $\E$ as before, the source of randomness in the sampling step, we can compute these node log-likelihoods without knowing the exact value for each $\varepsilon_k$. 

The setting with latent confounders ($\U$) is more complex. We do not have values for these confounders (they are non-observable), hence we cannot compute the previous formula directly. Specifically, if $U_{\{k, l\}} \in \U$, then $U_{\{k, l\}}$ is a parent for both $V_k$ and $V_l$; each log-likelihood term for these nodes would require its value. This can be by-passed by using the law of total probability: $\log p(v) = \log \mathbb{E}_\U[p(v \mid \U)] = \log \mathbb{E}_\U[\sum_{k=1..K} p(v_k \mid pa_k, U_{\{k, .\}})]$. This expectation can be approximated by Monte Carlo, taking $M$ i.i.d. samples from $\U$. Additionally, for numerical stability, we compute the individual likelihoods using log-likelihoods and then use the log-sum-exp trick to compute the log-expectation. This method allows the inclusion of latent confounders to our DCG models.

Notice that if the DCU's $\textbf{\textit{loglk}}$ operation is differentiable for all nodes, then the graph $\textbf{\textit{loglk}}$ is also differentiable. As a result, we can train all nodes in a graph simultaneously by Maximum Likelihood Estimation. This is a definite advantage in comparison with CausalGAN \cite{kocaoglu2017causalgan}, which used an adversarial setup, with significant drawbacks in harder training and possible mode collapse. It is for this reason that we require any DCU's $\textbf{\textit{loglk}}$ operation to be differentiable.

The last and final operation is \textbf{\textit{counterfactual}}, which generates $N$ counterfactual samples for all variables in $\V$ given a base sample $v$ and an intervention $\textrm{do}(X=x)$. These samples can be used to compute any expectations in the counterfactual model. In order to do that, we need to follow the three-step process defined before. The first one, \textbf{abduction}, can be performed by calling each node's $\textbf{\textit{abduct}}$ operation with its value and its parent's values. Depending on the implementation of each node's DCU, we are able to compute and sample from this posterior. When that is not possible, such as with discrete DCUs, we can still use rejection sampling: sample values for $E_X$ until the desired value $x$ is attained. Finally, if there are latent confounders, although rejection sampling is still possible, each $U_{\{k, l\}}$ now affects two nodes $V_k$, $V_l$, which increases the number of samples required for rejection sampling to work. In that case, we use \textbf{importance sampling}. Let us denote our counterfactual operation $CF_f\left(v, \textrm{do}(X=x)\right) := \expectation[(\E, \U \mid v)]{ f(V'(\E, \U, \textrm{do}(X=x))) }$ where $V'$ are the counterfactual samples and $f$ is a function of these samples. Then, $CF_f\left(v, \textrm{do}(X=x)\right) =$ $\expectation[\U]{ \expectation[(\E \mid v, \U)]{ f(V'(\E, \U, \textrm{do}(X=x))) } \, \frac{p(v \mid \U)}{p(v)} } \approx$ $\sum_{j=1..M} \expectation[(\E \mid v, u_j)]{ f(V'(\E, u_j, \textrm{do}(X=x))) } \, s(\log p(v \mid u))_j$, where $(u_j)_j$ are $M$ i.i.d. samples from $\U$ and $s(.)$ is the softmax operation. The derivation of this formula can be found in the supplementary material.

\subsection{Normalizing Causal Flows}

DCGs allow sampling, computing log-likelihoods of samples, perform interventions and compute counterfactual queries, provided an appropriate implementation of the DCU is given. Distributional Causal Nodes (DCN) \cite{parafita2019explaining} fulfill these requirements; they assign a parametric probability distribution to every node $V_k$ (\ie Gaussians, Exponentials or Categoricals) with parameters $\Theta_k$ and model their Conditional Probability Distribution (CPD) by defining a neural network $f_k$ that takes the node's parents as input and computes the distribution's parameters $\Theta_k$ as the output ($\Theta_k=f(Pa_k, U_{\{k, .\}})$). Note that we can perform all three DCU operations: 1) \textit{sample}, by using $E_X$ as an independent noise signal with a reparametrization trick \cite{kingma2013vae} for the assumed distribution; 2) \textit{loglk}, by using the density of the assumed parametric distribution; 3) \textit{abduction}, by inverting the reparametrization formula. This inversion is not always possible, in which case we can use rejection sampling, as discussed before (\ie for Bernoulli and Categorical distributions, when using the Gumbel trick to sample). 

There are, however, two disadvantages to DCNs. On the one hand, users need to specify a well-matched distribution for each node in the graph, which can be costly on graphs with many variables. On the other hand, standard distributions might not be sufficient to properly adjust complex datasets. Even though the marginal distribution of a DCN node is actually a (possibly uncountable) mixture of the assumed distribution, its CPD is still the base assumed distribution, which might be too restrictive in some datasets.

To avoid these two issues, we propose Normalizing Causal Flows (NCF), an alternative implementation of continuous DCUs. A Normalizing Flow models probability distributions by transforming a continuous random variable $X$ into a base distribution $E_X$, usually a standard normal distribution of the same dimension as $X$. This transformation is carried out by an invertible function $f$, which is guaranteed to exist under certain regularizing conditions \cite{papamakarios2019normalizing}. The main advantage of this setup is that we can compute $p_X(x) = p_{E_X}(f(x)) \cdot |\det J_f(x)|$, where $J_f(x)$ is the Jacobian of $f$ on $x$. Normally, we model this $f$ using a neural network with certain architectural constraints, so that it is invertible and this determinant is tractable. As a result, our flow $f$ is capable of: 1) \textbf{\textit{sampling}} from $X$ by taking an $\epsilon \sim p(E_X)$ and transforming it back to $X$ with $x = f^{-1}(\epsilon)$; 2) computing the \textbf{\textit{log-likelihood}} of a realization $x$ as described before; 3) computing the $\epsilon_x \sim p(E_X)$ such that $f^{-1}(\epsilon_x) = x$ (\textbf{\textit{abduction}}). 

\begin{figure}
\centering
\includegraphics{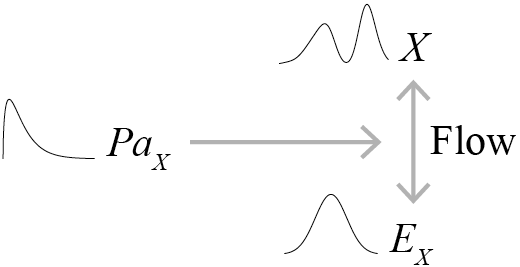}
\caption{Normalizing Causal Flow (NCF) component relationships. Flow transforms $X$ into its exogenous noise signal $E_X$, while $Pa_X$ acts as an input of the flow's conditioner. Consequently, NCF models the conditional distribution. }
\label{fig:flow_diagram}
\end{figure}

However, the distribution that we want to model is actually the CPD $(X \mid Pa_X, U_{\{X, .\}})$. Therefore, we need a Conditional Normalizing Flow: by adding the parent's values as an additional input of the flow's conditioner (more details on the transformer-conditioner framework in \cite{papamakarios2019normalizing}), we are effectively modelling the CPD and, as such, the three required operations of DCUs. With this model, any type of conditional Normalizing Flow can be used in a graph to model continuous random variables, \textbf{avoiding DCN's node-wise distributional assumptions}. Additionally, the high expressiveness of Normalizing Flows, as documented by innumerable works on the field, allow us to model much more complex distributions than with DCNs, as we will prove in the experiments. Figure \ref{fig:flow_diagram} summarizes the NCF unit, defined as the network $f$ that transforms $X$ into the exogenous noise signal $E_X$, by using the parameters returned by the conditioner network, that takes $Pa_X$ (and $\U_{\{X, .\}}$, if any) as its input. 

\subsection{Practical considerations}
\label{subsec:practical}

To speed up and stabilize training, it is very important to perform a \textit{warm start} of each node's distribution. For continuous nodes, we do this by learning normalization parameters based on the training data statistics. We perform normalization of a node's sample before computing its log-likelihood, and denormalize after sampling from that node's distribution. That way, the node learns how to model a distribution with location 0 and variance 1, which results in a more stable training. Note that the resulting log-likelihood also needs to take into account the normalizing transformation.

Regarding latent confounders, we model them as non-learnable nodes that just output an independent noise signal (usually, a univariate standard normal distribution). This signal is passed to its children as any other parent value and, by using the previously derived formula for the log-likelihood of a graph's sample, using Monte Carlo, we can train the whole graph so that it learns the effect of this latent confounder. In that regard, we found that a 100 samples for the Monte Carlo estimator is effective enough to learn the desired dependency.

\section{Experiments}
\label{sec:experiments}

In order to evaluate the use of our technique for SEM, we will consider several Deep Causal Graphs endowed with different DCU implementations. Since this is a probability distribution fitting problem and DCGs allow the computation of the mean log-likelihood of the dataset, we will compare different models by whoever maximizes this metric, or, equivalently, who \textbf{minimizes the negative log-likelihood (\textit{nll}) loss} of the test dataset. We will test our approach on the most popular dataset for each of the three following categories in the UCI Machine Learning repository \cite{dua2017uci_repo}: business (\textit{wine-quality}), life sciences (\textit{abalone}) and physical sciences (\textit{concrete}). Additionally, we also test it on a synthetic dataset, \textit{salary}, for which we know its causal graph. More details on this dataset can be found in the supplementary material.

For the sake of the evaluation, all three UCI datasets are modelled with a complete graph (every node depends on the previous nodes), except for the synthetic one, \textit{salary}, since we know its real causal graph. All discrete variables are modelled with Bernoulli or Categorical DCNs. Continuous variables, on the other hand, differ on the use of model: Generalized Linear Model (\textit{GLM}) assumes a Normal distribution and uses 1-layer networks (no hidden layers, effectively, linear models) for its parameter functions; Normal (\textit{Norm}) and Asymmetric Laplacian Distribution (\textit{ALD}) DCNs use 3-layer networks; Normalizing Causal Flow (\textit{Flow}) uses a conditional Deep Sigmoidal Flow \cite{huang2018dsf}.

\begin{figure}
\centering
\includegraphics[width=\linewidth]{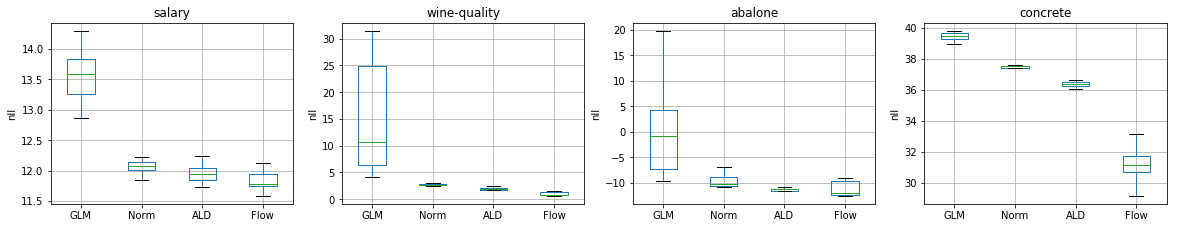}
\caption{Negative log-likelihood (\textit{nll}) boxplot of each model.}
\label{fig:nll_boxplots}
\end{figure}

Each dataset is divided in 10 splits for cross-validation, to obtain 10 estimates of the test \textit{nll} loss. Figure \ref{fig:nll_boxplots} contains the boxplots of these metrics. Note that \textit{GLM} has the poorest results in comparison with all Deep Learning models, which proves the advantage of DCGs over the commonly used approaches. On the other hand, Flows are one of the best options, if not the best, for all datasets. We also tested the adjustment for each variable, and Flows still emerged as the best model overall. We leave these plots for the supplementary material due to space restrictions.

\subsection{Sanity checks}
\label{subsec:sanity_checks}

\begin{figure}
\centering
\includegraphics[width=.9\linewidth]{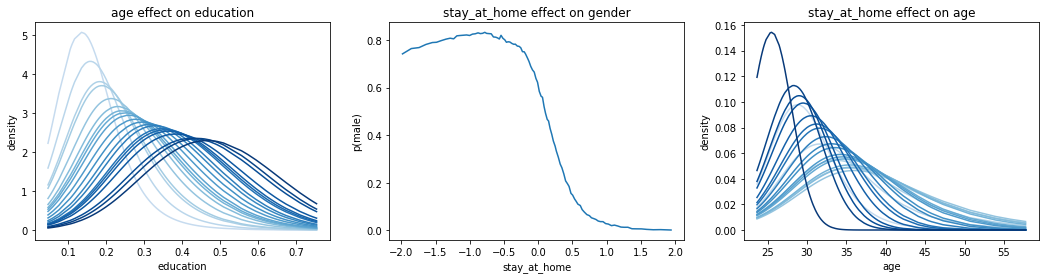}
\caption{Causal effect of a parent intervention on a node's distribution.}
\label{fig:conditional_effects}
\end{figure}

To test that our method really adjusts the generative process, we perform several sanity checks included in the code in the supplementary material. For one, we can test the marginal distribution of each node in the graph to see if it fits with that in the data, but we also need to \textbf{confirm that the Conditional Probability Distribution is properly modelled}. If the parent's values did not have an effect on the node's density, we would be adjusting only the marginals, which would not model interventions as desired. For that reason, we plot each node's density conditioned on interventions on its ancestors. Figure \ref{fig:conditional_effects} shows three of these experiments for the synthetic \textit{salary} dataset. 

This dataset was designed to represent societal biases against women in the workplace so that we can test a predictor's fairness with respect to a certain group, as we will see in the following section. The left plot in figure \ref{fig:conditional_effects} shows the effect of \textit{age} on the \textit{education} level of that person. Here we plot the density function of \textit{education} subject to interventions on \textit{age}, where darker lines represent higher values of \textit{age}. This tells us that older people are more likely to have higher values of \textit{education}, given that their density is progressively shifting to the right. Indeed, the node captures the CPD as desired.

On the other hand, we also confirm that \textbf{DCGs are capable of learning latent confounder effects}. In this dataset, we introduced selection bias by filtering women stochastically with an increasing probability as they get older. This is used to represent stay-at-home mothers: since these women would not be receiving a salary, they would not appear in the dataset. By filtering them out, we are implicitly introducing a latent confounder (\textit{stay at home}) between \textit{gender} and \textit{age}. Had the DCG captured this relationship, we would expect the distribution of these nodes to be affected by the confounder. Indeed, the center plot in figure \ref{fig:conditional_effects} shows the dependency between the probability of being a male ($p(\textit{male})$) and the values of the confounder, while the right plot shows the density curves for \textit{age} with different values for \textit{stay at home}. Finally, the correlation between \textit{gender} and \textit{age} in a sample from the trained graph is 29\%, the same as in the original dataset. This, again, shows that the graph has learnt the dependency caused by the latent confounder.

\section{Counterfactual Explainability and Fairness}
\label{sec:applications}

This section is devoted to showcasing applications of DCGs to Counterfactual Explainability and Fairness, using the aforementioned \textit{salary} dataset. We start by training a NCF causal model and a Neural Network regressor for \textit{salary}, that will be considered our black-box predictor. The objective is to explain a certain prediction on an individual (\textbf{explainability}), to determine whether the predictor is counterfactually fair \cite{kusner2017counterfactual}, and to re-train it so that it is indeed fair (\textbf{fairness}). Details on the dataset, its causal graph and further experiments on \textbf{interpretability} and on the use of continuous variables for interventions are left for the supplementary material due to space restrictions.

Looking at the dataset, we notice that there is a clear bias in average \textit{salary} between women and men: \$27,734 to \$32,856. The black-box regressor also captures this bias. It looks like the data (and consequently, the model) is unfair towards women. We now look for a particular woman (sample 770, 0-indexed) who, due to her input features, we suspect her prediction was specially unfair: \$31,548. What we want to measure is, had this woman been a man, what would her \textit{salary} be? To answer this query, we need to use counterfactuals: get the \textbf{abducted} noise for this sample, apply the desired \textbf{intervention} ($\textrm{do}(\textrm{gender} = \texttt{male})$) and obtain multiple \textbf{counterfactual samples}. We pass each sample through the regressor and aggregate all counterfactual predictions; the result is the counterfactual estimation: \$37,586. Indeed, it would be particularly unfair to use the predictor with this woman.

Although we found an unfair sample, this might not be the case with the whole population. Is the regressor (and the data from which it learnt) fair \wrt to the protected variable (gender)? Counterfactual Fairness \cite{kusner2017counterfactual} essentially computes the previous expression for the whole population. We define Counterfactual Unfairness of degree $k$, $CU_k$, as $CU_k := \expectation[\V]{\expectation[\E, \U \mid \V]{|Y'(\textrm{do}(X=x), \E, \U)  - Y(\E, \U)|^k}}$ where $X$ are the intervened (protected) variables, $Y$ the target variable and $Y'$ the counterfactual target variable. The $CU_1$ of \textit{salary} (the average unsigned difference between counterfactual and real values) is \$4,218, which shows a clear bias in the model. Note that \textbf{we can train the model adding $CU_2$ as a regularization term}; the resulting predictor has a $CU_1$ of \$924, making it, indeed, fairer.

However, aligning these two populations would create discrepancies with the original dataset, which should reduce our performance. Note that by adding $CF_2$ as a regularization term, \textbf{we are making the predictor learn from a non-biased distribution, different from the observed distribution entailed by the dataset}. This means that low performance with respect to the original data does not mean poor performance by the model. For this case, we might be more interested in ensuring that the resulting ranking in salary is preserved by the model inside each group defined by our protected variable. We can measure this with the Spearman correlation: for the original model, it was 93\% for women and 89\% for males, while for the fair model, 84\% and 78\% respectively. In summary, when evaluating CU-regularized models, it is important to consider which metric is adequate to measure performance, since the original data might not reflect the fair world we want to model.

Finally, we can test this new fair regressor on the previous woman: the predicted and counterfactual \textit{salary} estimations become \$31,548 for the original woman and \$32,315 for the counterfactual male, with a difference of \$767 in contrast with the \$6,038 gap from the original model, thereby reducing the gender gap. In summary, \textbf{we can use Counterfactual Fairness, estimated with DCGs, to train black-boxes to be more fair through regularization}.

\section{Further work}
\label{sec:further}

DCUs are an abstract specification of the functionality required by a causal graph to run causal queries. As such, \textbf{alternative implementations of DCUs} could be tackled. For example, we tested DCNs using compound distributions with moderate success. Compound distributions are parametric distributions where its parameters are themselves random variables with learnable priors; these are modelled by adding an additional exogenous noise signal to the parameters network input. Further work on this approach or other types of DCUs could be tackled.

On the other hand, in this work we modelled univariate random variables exclusively. In theory, NCFs should adapt to more dimensions seamlessly, but further experiments on this problem remain for future work. Additionally, another interesting question is how to adapt DCGs to non-i.i.d. data, like time-series. For this case, we might add feedback loops to our graphs, to model the effect of the past on the future, working on the same set of variables.

\section{Conclusions}
\label{sec:conclusion}

We propose Deep Causal Graphs, a general technique for applying Neural Networks to the field of Causal Modelling. This framework allows modelling complex data distributions without sacrificing counterfactual estimation. Additionally, latent confounders can also be modelled with our technique. 

We provide a flexible model for the graph's unit, Normalizing Causal Flows, and demonstrate its fitting capabilities with our experiments. We also propose applications of our techniques to the fields of black-box explainability and counterfactual fairness, using true causal counterfactuals. Finally, we provide a complete software library, in the supplementary material and soon as an open-source project, for anyone to model their data using DCGs and test any of these applications.

\clearpage
\section*{Broader Impact}
\label{sec:impact}
Causal modelling is a general technique that allows us to study the generative mechanisms of the data. We propose a novel method to answer causal queries on datasets with complex relationships between variables, and provide a software library to use this technique. \textbf{We expect our research to act as a vehicle for the democratization of causal inference applications.}

In that sense, causal inference can be employed to estimate the causal effect of an intervention on an outcome variable, with applications to epidemiology, economics or business, to name a few. It can also be used to interpret black-box predictors, explain particular outcomes or assess and train towards counterfactual fairness. In both regards, the contribution of our system to society is fundamentally positive. It also allows black-box systems to be in compliance to the GDPR requirement of transparency whenever an automated decision might affect individuals.

There are, however, some risks to take into account. Since causal estimations depend on the graph structure, the validity of any explanation is contingent on the \textit{causal sufficiency} \cite[chapter~9]{peters2017elements} of the structural model with respect to the real data-generating process. By ignoring the effect of latent confounders, we might obtain contradictory results alike the ones in the well-known Simpson's Paradox, where a treatment effect can be both positive and negative depending if we condition on a group or not. This result, more than a paradox, is an effect of not considering the appropriate causal structure. As such, \textbf{great care is required in defining the graph before attempting to draw conclusions from it}.

In line with this reasoning, \textbf{malicious actors might use wrong causal graphs in an attempt to conceal their algorithmic biases}. In that sense, causal explanations might excuse unfair systems. Fortunately, since our approach allow us to inspect the effect of variables on any black-box system, as long as we have access to its input-output routine we are capable of detecting these biases. It is for this reason that we cannot find any negative repercussions that cannot be countered by our same methodology, provided we can derive an appropriate causal graph. 

\section*{{\small Acknowledgements}}
\noindent {\small This work has been partially funded by projects TIN2015-66951-C2, RTI2018-095232-B-C21 (MINECO/FEDER) and 2017.SGR.1742 (Generalitat de Catalunya).}

\bibliographystyle{unsrt}
\bibliography{Causal_Inference_with_Deep_Causal_Graphs}

\clearpage
\appendix

\section{Importance sampling}
\label{sec:importance_sampling}
When dealing with latent confounders, one must use importance sampling to properly deal with counterfactual estimations. Let us denote our counterfactual operation $CF_f\left(v, \textrm{do}(X=x)\right) := \expectation[(\E, \U \mid v)]{ f(V'(\E, \U, \textrm{do}(X=x))) }$ where $V'$ are the counterfactual samples and $f$ is a function of these samples. We can apply importance sampling on $\U$ in the following way (assuming continuous random variables with densities $p$):

\begin{equation} 
\begin{aligned}
    CF_f\left(v, \textrm{do}(X=x)\right) :=\, & \expectation[(\E, \U \mid v)]{ f(V'(\E, \U, \textrm{do}(X=x))) } = \\
        =\, & \expectation[(\U \mid v)]{ \expectation[(\E \mid v, \U)]{ f(V'(\E, \U, \textrm{do}(X=x))) } } = \\
        =\, & \int \expectation[(\E \mid v, u)]{ f(V'(\E, u, \textrm{do}(X=x))) } \, p(u \mid v) \, du  = \\
        =\, & \int \expectation[(\E \mid v, u)]{ f(V'(\E, u, \textrm{do}(X=x))) } \, \frac{p(v \mid u) \, p(u)}{p(v)} \, du  = \\
        =\, & \expectation[\U]{ \expectation[(\E \mid v, \U)]{ f(V'(\E, \U, \textrm{do}(X=x))) } \, \frac{p(v \mid \U)}{p(v)} } \approx \\
        \approx\, & \sum_{j=1..M} \expectation[(\E \mid v, u_j)]{ f(V'(\E, u_j, \textrm{do}(X=x))) } \, s(\log p(v \mid u))_j
\end{aligned}
\label{eq:importance_sampling}
\end{equation}

where $u = (u_1, \dots, u_M) \sim p(\Proba_\U)$ are $M$ i.i.d. samples from $\U$ and $s(.)$ is the softmax operation. This last step comes from the realization that $p(v) \approx \frac{1}{M} \sum_{j=1..M} \exp \log p(v \mid u_j)$ and, therefore, $\frac{p(v \mid u_j)}{p(v)} \approx \frac{\exp \log p(v \mid u_j)}{\frac{1}{M} \sum_{j=1..M} \exp \log p(v \mid u_j)} = M \cdot s(\log p(v \mid u))_j$. The $M$ term is cancelled by the expectation-average denominator, rendering the formula described by equation \ref{eq:importance_sampling}.

\section{Applications}
\label{sec:applications_sup}

We devote this section to describing the reasoning behind the \textit{salary} dataset and including further experiments showcasing applications of DCGs to interpretability and explainability. The implementation of these experiments is also included in the supplementary material.

\subsection{Salary dataset}
\label{subsec:salary_dataset}

The objective in creating the \textit{salary} dataset was to evaluate interventions and counterfactuals in a clearly defined context, so that the conclusions derived from our method could be assessed properly. The majority of real life datasets do not come together with a causal graph, and even if we derive it with structure learning algorithms and prior knowledge, their structure might not suffice for our purposes. The \textit{salary} dataset was constructed so we could evaluate discrete and continuous interventions and counterfactuals, even in the presence of latent confounders, so as to test every possible application of our technique.

\begin{figure}
\centering
\includegraphics[width=.5\linewidth]{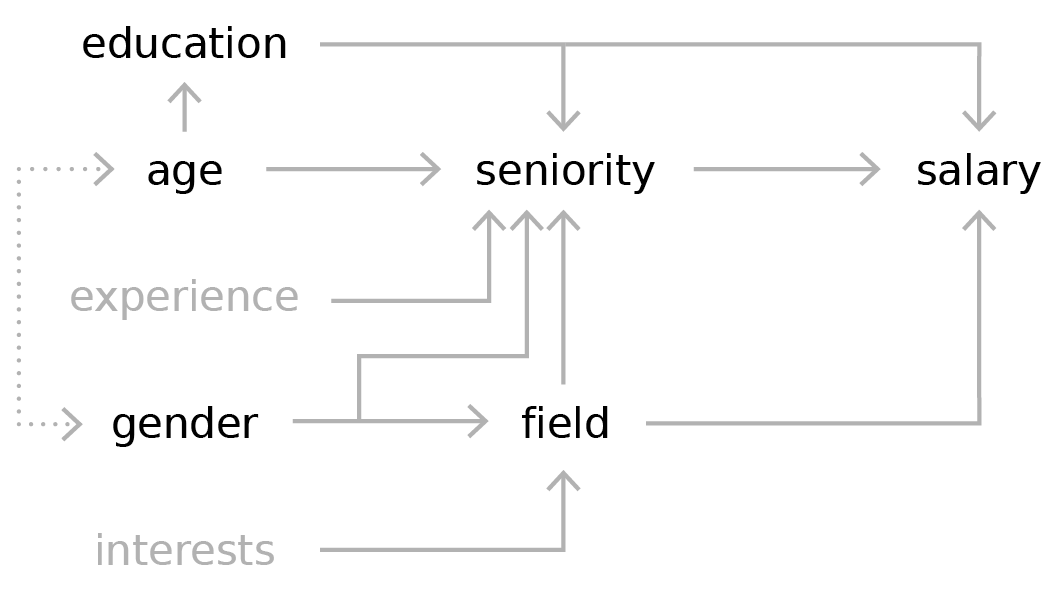}
\caption{Causal model for the \textit{salary} dataset. Grey variables are latent.}
\label{fig:salary_graph}
\end{figure}

Figure \ref{fig:salary_graph} shows the causal graph employed in the construction of each sample. Here, \textit{interests} and \textit{experience} represent two latent variables (non-observable, not included in the dataset) that affect the values of their respective descendant variables. We include them to assess the ability of our exogenous noise signals of capturing this dependency. They are also useful to test counterfactual estimation: if we study an individual with high experience (therefore, higher seniority) we expect its counterfactuals to capture this value implicitly (abduction). All variables are continuous, except for \textit{gender} and \textit{field}, which are modelled as Bernoulli distributions. Finally, the dashed bidirectional arrow between \textit{age} and \textit{gender} reflects a latent confounder resulting from selection bias, as discussed in the paper.

Since this dataset is used to simulate gender biases on the workplace, we will encode some causal relationships that represent this bias. As an example, gender affects the job field in which a person works, presumably as a result of societal pressures on the choice of field. Gender also affects seniority, the level of responsibility and power inside the company; this represents the bias against women towards better paid positions. 

The variable of interest is \textit{salary}, which we expect to be unfair towards women. We will study the effect of interventions of gender on a regressor that has been trained to predict that variable. Note that this can be modelled as another graph, where we replace the \textit{salary} variable by a new \textit{output} node that has every other node as a parent. This means, the value of all input variables for the regressor affects its output, but these variables are also related in a causal way as described by the original graph. The regressor itself is the functional form of the relationship between its output and its input variables, and we are interested on interventions on some input variables and how they affect the regressor's prediction. This is what we will study in the following subsections.

\subsection{Interpretability}

We are interested in two variables and their effects on \textit{salary}: \textit{gender} and \textit{age}, discrete and continuous. We can compute the effect of interventions on these variables to \textit{salary}. For that, we generate 1,000 samples from each intervened graph (1,000 for each possible intervention) and pass them through our regressor to obtain the predicted \textit{salary}. We intervene on both values for \textit{gender}, and compute a confidence interval for their means (at 95\% confidence), with [\$26,870, \$27,461] for women and [\$31,243, \$31,869] for men. Note that these intervals could be different to the ones obtained from conditioning on the original dataset, since \textit{gender} is not a root node and, therefore, the observational model and the gender-intervened model are not equivalent when conditioning on gender. Indeed, [\$26,094, \$26,627] for women and [\$32,761, \$33,407] for males. These differences appear due to the effect of the conditioning value of \textit{gender} on \textit{age}; the latent confounder connects them, altering the distribution for age and therefore unduly perturbing the results (since by changing \textit{gender} we should not be altering \textit{age}). To solve this discrepancy we could use back-door adjustments, but our method avoids this by using the structural model directly.

\begin{figure}
    \centering
    \includegraphics[width=.45\linewidth]{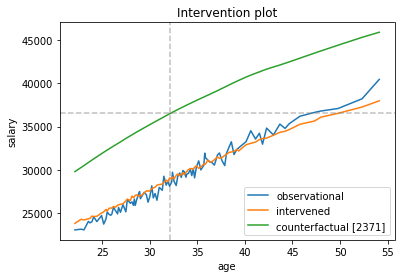}
    \includegraphics[width=.45\linewidth]{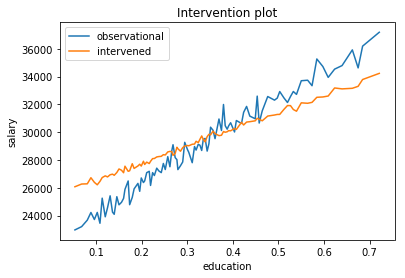}
    \caption{Intervention plots for age and education.}
    \label{fig:intervention_plots}
\end{figure}

We can test this same technique with continuous variables. Here, we generate 100 equidistant values between 0.025 and 0.975 and compute their corresponding quantiles for the \textit{age} and \textit{education} variables. For each of these values, we intervene with them on the graph and compute 1,000 samples, which we aggregate and plot in figure \ref{fig:intervention_plots}. We also compute the mean salary for all samples whose \textit{age} or \textit{education} falls in one of these quantiles and plot them as the \textit{observational} curve. In this case, the difference between the observational and interventional curves for \textit{age} is not that noticeable, but it is for \textit{education}, since \textit{education} is correlated with \textit{age} and by conditioning on the latter, we alter the prior on the former, which is not possible in the intervened graph. The counterfactual curve in the \textit{age} plot will be discussed in the following subsection.

This type of experiment allow us to study the effect of the intervened variables on the final prediction. Were we to ask someone to change their field, irrespective of their interests and gender pressures, we would expect a mean difference in salary of \$5,761. Note that this intervention does not mean to just change the value in the \textit{field} column, but also take into account the effects of this alteration on its descendants, namely, \textit{seniority}. It is these effects that actually produce the expected change in \textit{salary}. Therefore, we can perform attribute-level \textbf{interpretability} of any black-box system by using interventions with our DCGs.

\subsection{Explainability}

In this section we are interested on explaining particular predictions on an individual. Specifically, we look for a woman whose \textit{salary} would increase significantly had she been a man, and a person with average age whose \textit{salary} would decrease significantly had they been younger. In other words, we look for counterfactual estimation with interventions on \textit{gender} and \textit{age}, respectively.

The first case, sample 770, 0-indexed, corresponds to the one discussed in the paper. Her expected \textit{salary} grows from \$31,548 to \$37,586. Note that an intervention on \textit{gender} can affect \textit{field}, changing it, from field A to B, in 89\% of the cases. This change results from societal pressures encoded by the dataset, where women are normally attracted to field A, in contrast with men. This change in \textit{gender} and \textit{field} also affects \textit{seniority}, the most relevant variable for the regressor, changing its value from 1.75 to 2.36. The rest of the input variables, since they are not causally connected with \textit{gender}, are not affected by the intervention.

For the second case, sample 2,371, 0-indexed, figure \ref{fig:intervention_plots} left shows the counterfactual effect of age (counterfactual). Dashed lines represent the real age and predicted salary for that sample. This individual has a noticeably higher value for \textit{salary} than its peers at their original \textit{age}, 32, and this fact is captured by the counterfactual (through abduction) across all possible interventions on age. This is mainly caused by their value on \textit{education}, 0.73, which is a very high value overall. At 22 years old, \textit{education} becomes, on average for all 1,000 counterfactuals, 0.43, a high value still, which eventually affects both \textit{seniority} (from 1.88 at the original age to 1.53 at 22 years old) and \textit{salary}.

\section{Variable fitting experiments}
\label{sec:variable_fitting}

In this section we include several plots showing the fitting capabilities of each of the tested methods with each continuous variable in the four datasets. The left plot is the boxplot of negative log-likelihood for each variable; the center plot is the real data histogram and the resulting KDE curves of samples generated by each of the models; the right plot is the Cumulative Distribution Function (CDF) plot of the original data and each model. As mentioned in the paper, Flows are among the best options in the majority of the variables, if not the best one.

\begin{figure}
\centering
\includegraphics[width=.9\linewidth]{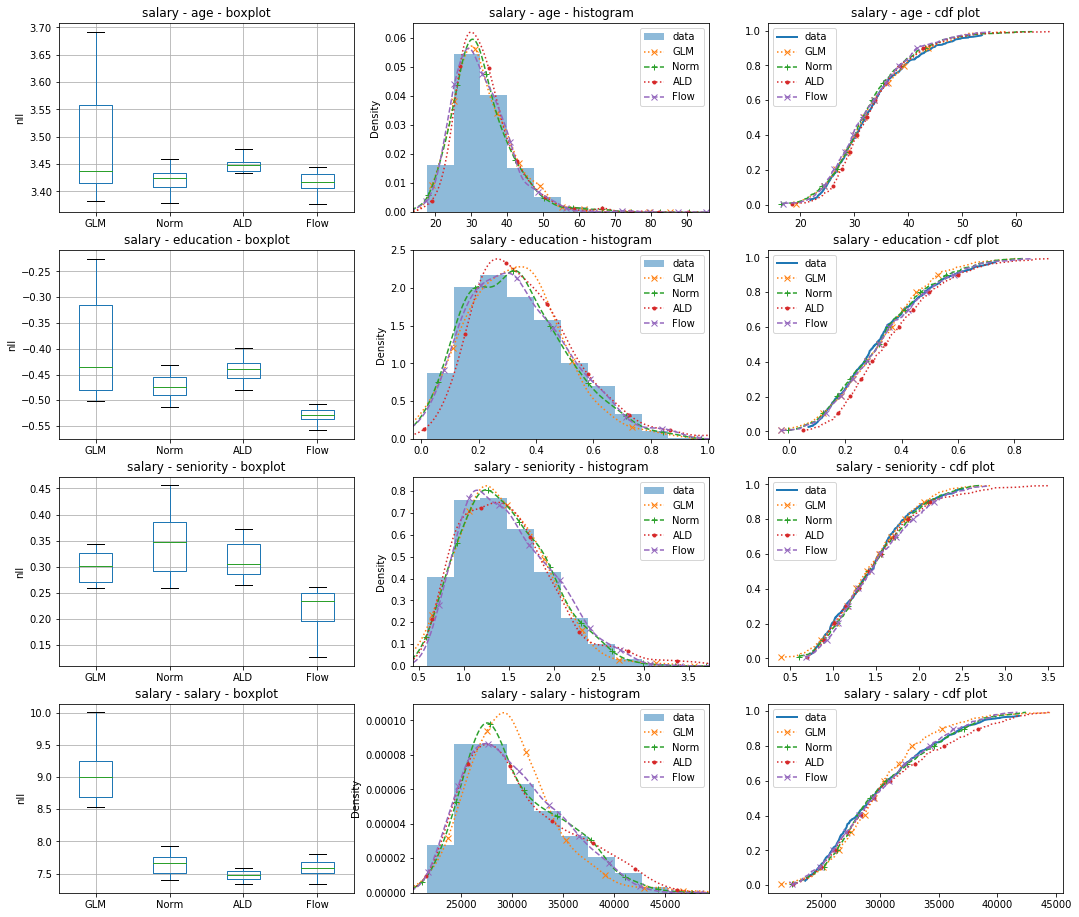}
\caption{Variable fitting experiments for the \textit{salary} dataset.}
\label{fig:variable_salary}
\end{figure}

\begin{figure}
\centering
\includegraphics[width=.7\linewidth]{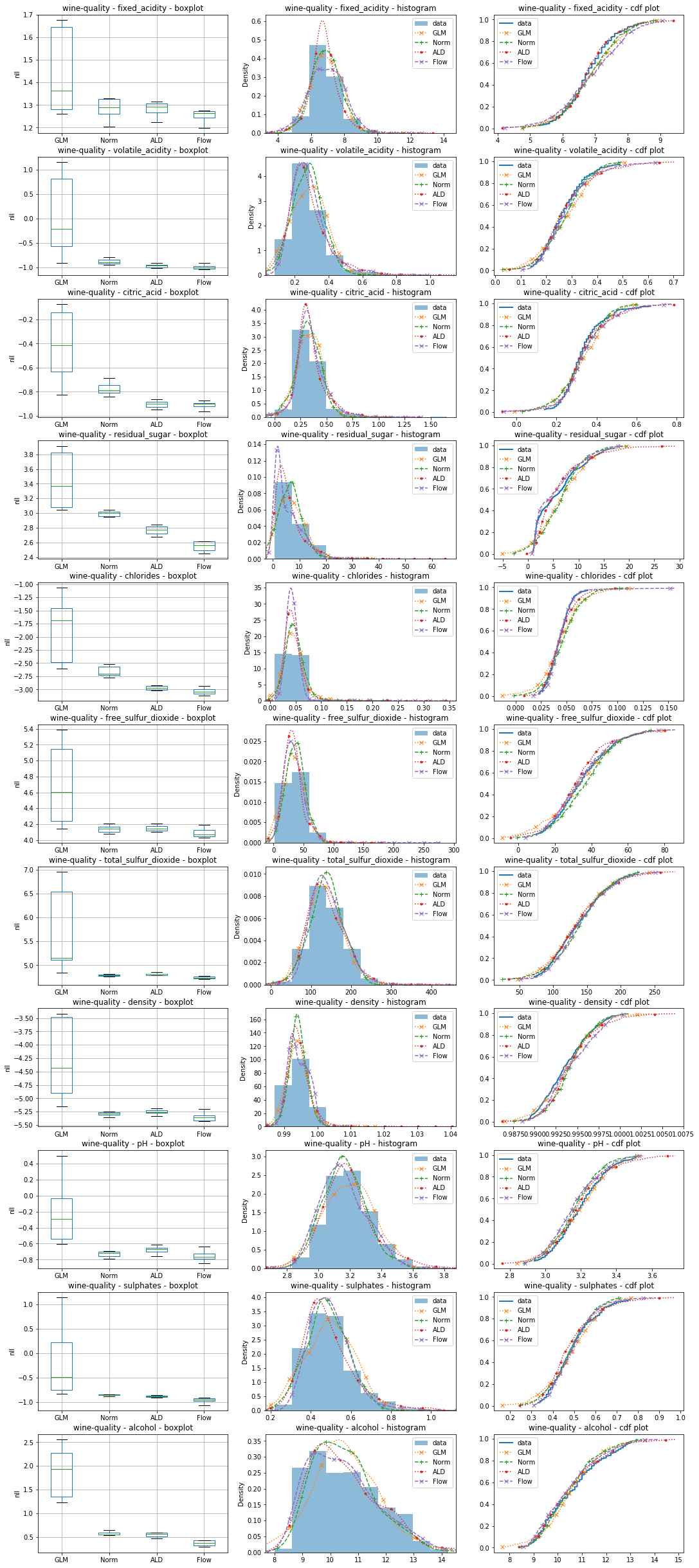}
\caption{Variable fitting experiments for the \textit{wine-quality} dataset.}
\label{fig:variable_wine}
\end{figure}

\begin{figure}
\centering
\includegraphics[width=.9\linewidth]{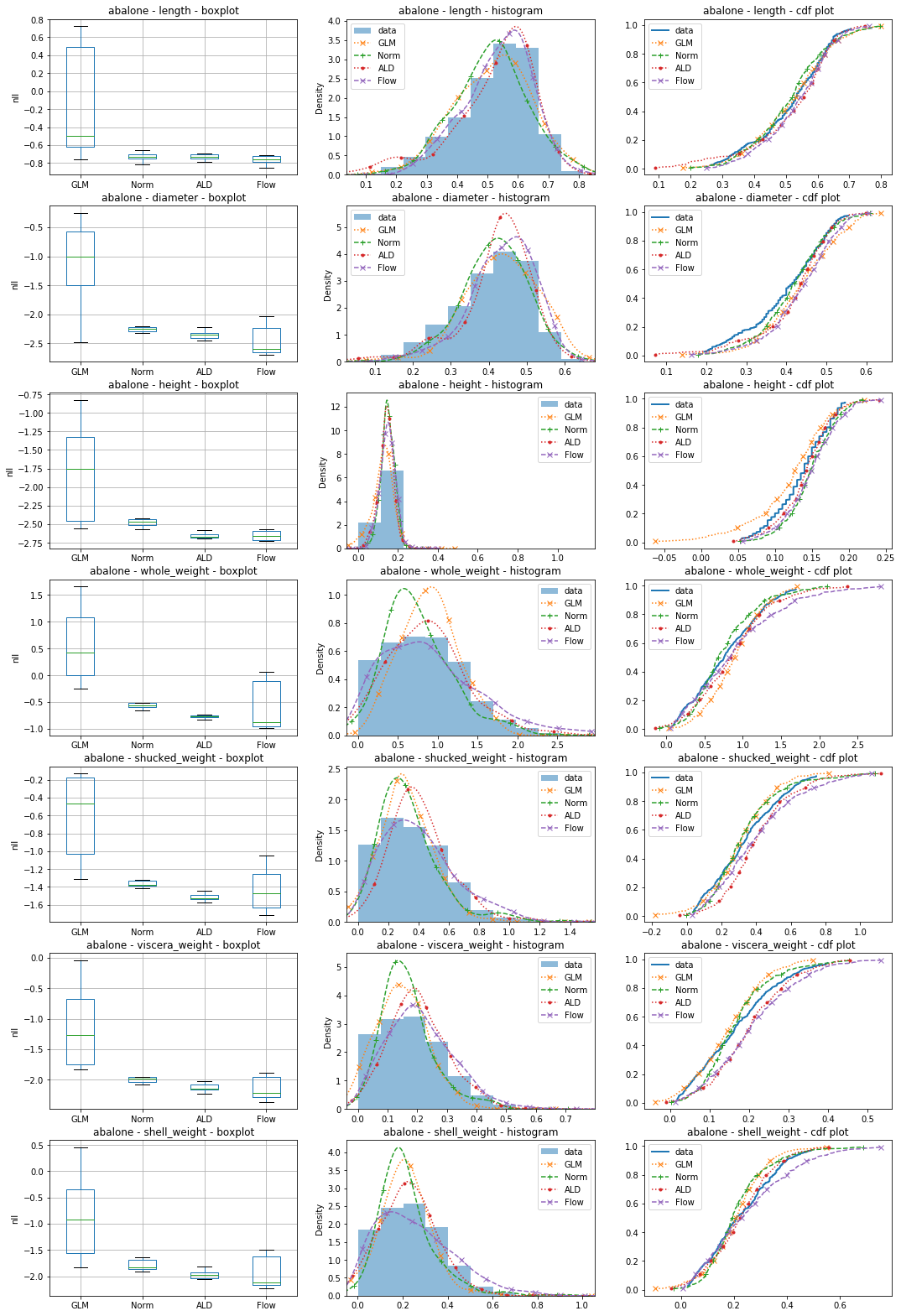}
\caption{Variable fitting experiments for the \textit{abalone} dataset.}
\label{fig:variable_abalone}
\end{figure}

\begin{figure}
\centering
\includegraphics[width=.9\linewidth]{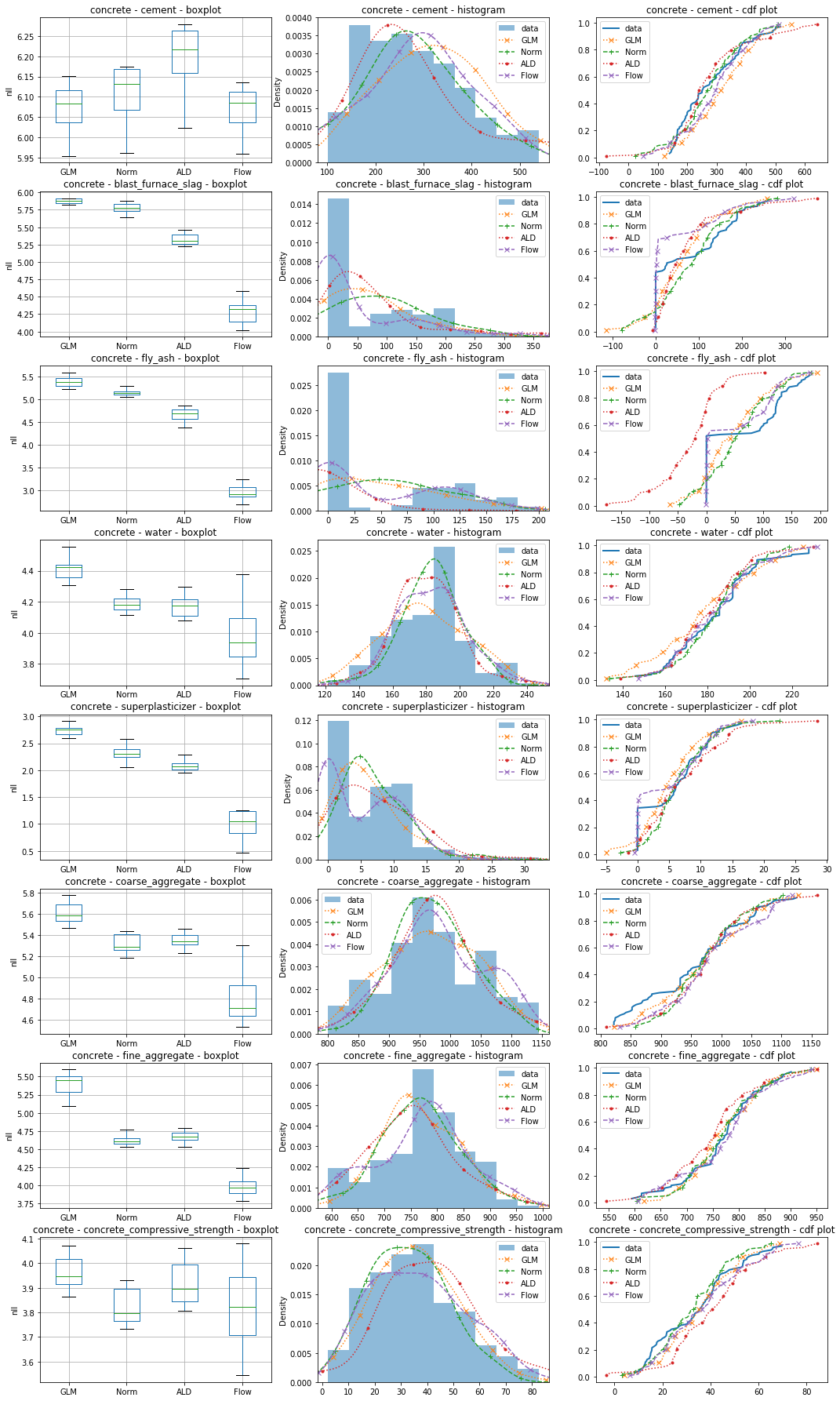}
\caption{Variable fitting experiments for the \textit{concrete} dataset.}
\label{fig:variable_concrete}
\end{figure}

\end{document}